%% file: main_iclr.tex
\documentclass{article} %
\usepackage{iclr2027_conference,times}

\input{math_commands.tex}

\usepackage{hyperref}
\usepackage{cleveref}
\usepackage{url}
\usepackage{graphicx}
\usepackage{wrapfig}
\usepackage{booktabs,multirow,graphicx,siunitx,colortbl}

\title{FurE: Efficient Instance-Specific 3D Fur Reconstruction without Animal-Fur Datasets}

\author{
Srinjay Sarkar$^{\star}$ \qquad
Prakhar Kaushik$^{\star\dagger}$ \qquad
Soumava Paul \qquad
Alan Yuille
\\[0.4em]
{\small
$^{\star}$ Equal contribution
\qquad
$^{\dagger}$ Project lead
}
\\[0.6em]
Johns Hopkins University
\\
Baltimore, MD, USA
\\
\texttt{\{ssarka29, pkaushi1, spaul27, ayuille1\}@jh.edu}
\\
\url{https://toshi2k2.github.io/fure}
}

\iclrfinalcopy %
\begin{document}
\maketitle
\footnotetext[1]{Work done while Srinjay was an intern at JHU.}
\begin{figure}[h]
    \centering
    \includegraphics[width=\textwidth]{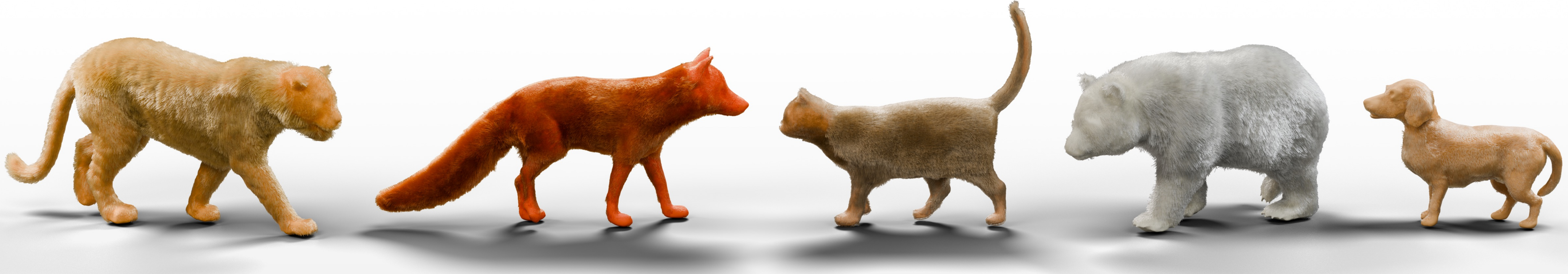}
    \caption{From multi-view images, FurE \textit{efficiently} reconstructs a defurred animal body and  editable, strand based fur geometry. We utilize fur thickness cues from learned volumetric surface representations
    to estimate fur depth and an estimate of the unseen defurred body.}
\end{figure}

\input{sec/0_abstract} 
\input{sec/1_intro}
\input{sec/2_Related_Work}
\input{sec/2_method_preprocess}

\input{sec/3_method}
\input{sec/4_experiments}
\input{sec/5_conclusion}

{
    \small
    \bibliographystyle{iclr2027_conference}
    \bibliography{main}
    
    \appendix
    \input{sec/appendix}

}

\end{document}

%% file: math_commands.tex
\usepackage{amsmath,amsfonts,bm}

\def\eqref#1{equation~\ref{#1}}

\def\1{\bm{1}}

\DeclareMathAlphabet{\mathsfit}{\encodingdefault}{\sfdefault}{m}{sl}
\SetMathAlphabet{\mathsfit}{bold}{\encodingdefault}{\sfdefault}{bx}{n}

%% file: sec/0_abstract.tex
\begin{abstract}
Realistic and editable animal fur reconstruction from multi-view images is challenging: fine-scale detail, self-occlusion and obfuscation, and, unlike human hair, the lack of animal fur datasets. Fur usually covers most of an animal's body, with large inter- and intra-species variability.
We present \textit{FurE}, an efficient strand-based animal fur reconstruction method that recovers a per-strand, editable groom by optimizing a root-conditioned latent field, decoded into strand geometry via a PCA-based decoder. 
We reconstruct a defurred animal body using local fur thickness cues from surface-constrained Gaussian "frosting" representation, along with part-based priors. We next show using a PCA-based decoder using knowledge from human hair strand data - allows us to alleviate the animal data scarcity, while allowing for faster optimization.
FurE achieves a 10$\times$ speedup in strand training over current SOTA dense per-strand optimization, retaining strand fidelity and generalizing across synthetic and, more importantly, real-world sequences, with quantitative and qualitative validation despite this reduction in training time. 
\end{abstract}

%% file: sec/1_intro.tex
\section{Introduction}
\label{sec:intro}
Strand-based hair and fur is the standard representation for high-quality digital assets. Unlike volumetric or surface-based representations, explicit strands remain directly editable, renderable, and simulatable in production pipelines. Recovering such a representation from images is difficult, especially for animal fur where individual fine-scale strands are heavily self-occluded by surrounding fur, and this occlusion is compounded by fur's view-dependent appearance and the fact that it covers most of the animal's visible body. Moreover, the surface reconstructed from multiview images is usually the outer furry envelope rather than the underlying skin on which the strand roots should be placed. This ambiguity is substantially greater than in human hair capture, as animal fur varies not only across species but also across body parts of the same individual, with different lengths, densities, directions, and thicknesses around the face, ears, torso, belly, legs, paws, and tail. Disentangling fur from the body also remains crucial for tasks such as pose estimation~\citep{xu2023animal3d}, 3D part segmentation, and tracking.

We present \textbf{FurE}, an efficient method for strand-based animal fur reconstruction from multiview images, built on two principles. First, the expensive fur strand-learning problem should be solved in a low dimensional latent space rather than the full 3D strand space. Instead of treating each fur strand as an unconstrained high-dimensional polyline, FurE predicts compact (low-dimensional) codec coefficients (vectors) on a defurred body surface, which a lightweight pretrained (using abundant human hair data) PERM~\citep{perm2025} style decoder maps to explicit local-frame strand geometry, scaled by fur-length and rendered with strand-aligned gaussians. This retains the editing and rendering benefits of strand-based reconstruction while substantially reducing the cost of per-scene strand learning relative to dense per-strand optimization used in current SOTA methods.

Second, FurE treats defurring (estimating the skin underlying the fur) as a local \textit{shell}-estimation problem (outer shell being the visible fur, and inner is defurred).  We discovered that we can use fur-depth cues from Gaussian Frosting~\citep{guedon2024gaussianfrosting}, which uses the misalignment of surface-aligned Gaussians to identify areas where more volumetric rendering is needed.

FurE extracts view-consistent shell statistics as local evidence for fur-bearing volume, then calibrates this evidence with part-aware priors to produce a defurred surface mesh. Given this defurred mesh approximation, FurE samples strand roots and assigns each a local coordinate framework - TBN basis (Tangent, Bitangent, Normal), semantic part label, and calibrated length; a latent UV texture is then decoded via a lightweight PCA decoder into normalized canonical strands, scaled to target length and transformed into world space. FurE then attaches cylindrical Gaussians to the decoded strand segments and optimizes with multiview photometric loss, yielding explicit polylines and strand-aligned Gaussians compatible with downstream rendering, simulation, and editing applications.

On multiview inputs from the Artemis dataset~\citep{10.1145/3528223.3530086}, FurE completes strand training in under one hour, achieving a $10\times$ speedup with comparable or better quantitative results while retaining explicit, editable strands. We evaluate reconstruction quality and efficiency against animal fur and strand-based hair baselines, with ablations of the codec representation. We further demonstrate, to our knowledge, the first instance-specific, strand-based animal fur reconstruction from noisy real-world multiview images, on which NeuralFur~\citep{NeuralFur26} struggles. 

Our contributions are:

\begin{itemize}
\item We introduce \textit{FurE} for instance-specific reconstruction of explicit, editable animal fur from calibrated multiview images. Strand training takes under one hour, achieving a $10\times$ speedup over SoTA methods with comparable or better rendering quality.

\item We transfer human-hair priors to animal fur through a compact PCA codec, replacing dense strand optimization with root-conditioned latent learning without animal-fur training data.

\item We introduce shell-based defurring that combines local Gaussian Frosting cues with part-aware priors to estimate a plausible hidden strand-root surface without SMAL fitting.

\item We demonstrate, to our knowledge, the first instance-specific, strand-based animal fur reconstruction from noisy real-world multiview images.
\end{itemize}

%% file: sec/2_Related_Work.tex
\section{Related Work}
\label{sec:related_works}

\textbf{Animal reconstruction.}
SMAL~\citep{Zuffi:CVPR:2017}, GenZoo~\citep{niewiadomski2024generativezoo}, and AniMer~\citep{lyu2025animer} recover animal body shape but not explicit fur strands. NeuralFur, our closest prior work, combines NeuS geometry~\citep{wang2021neus}, SMAL-based part localization, and VLM-derived fur attributes to guide defurring and strand reconstruction. Its defurring depends on template fitting and part-level semantic estimates, while dense strand optimization remains computationally expensive. FurE instead uses local Gaussian Frosting cues and part-aware priors to estimate the hidden strand-root surface. Direct part segmentation~\citep{alignparts2025} removes the dependency on SMAL fitting. Combined with compact strand learning, this achieves a $10\times$ strand-training speedup with comparable rendering quality. We further reconstruct explicit fur from a noisy real-world bison sequence on which NeuralFur fails.

\textbf{Strand-based and compact hair reconstruction.}
Neural Haircut~\citep{sklyarova2023neural_haircut} introduced prior-guided strand reconstruction. Gaussian Haircut~\citep{zakharov2024gaussianhaircut} and GaussianHair~\citep{Luo2024GaussianHairHM} use strand-aligned Gaussians for differentiable rendering, while CGHair~\citep{CGHair2026} reduces memory through strand/card clustering and shared appearance codes. PERM~\citep{perm2025} represents human hair using compact PCA coefficients, while GroomGen~\citep{zhou2023groomgen} uses hierarchical latent representations. FurE transfers a human-hair PCA prior to instance-specific animal fur reconstruction, optimizing root-conditioned latent codes rather than dense strand geometry. This reduces per-scene optimization cost without requiring animal-fur training data, while retaining explicit, editable strands.

%% file: sec/2_method_preprocess.tex
\input{figures/method_fig}
\section{Method}

Given calibrated multiview images, FurE first estimates a defurred body mesh $M_{\mathrm{root}}$ and a target strand length $\ell_p$ for each body part $p$ (\cref{sec:frosting_preprocessing}). We sample roots $\mathbf{r}_i$ (strand attachment points) on this mesh and predict a compact vector of shape coefficients $z_i$ from each root's position, part label, local thickness cue, and target length. A PCA decoder $D$, learned from human-hair data, converts these coefficients into a local strand shape, which is scaled to $\ell_p$ and oriented and positioned at the root (\cref{strand_gen}). We optimize this latent representation against the input views through differentiable rendering with strand-aligned cylindrical 3D Gaussians (\cref{render_opt}).

To estimate $M_{\mathrm{root}}$, we move each vertex $\mathbf{x}_i$ of the visible furry mesh $M_{\mathrm{outer}}$ inward along its outward normal $\mathbf{n}_i$:
$
\mathbf{x}_i^{\mathrm{root}} = \mathbf{x}_i - d_i \mathbf{n}_i,
$
where $d_i$ is the estimated inward displacement. We infer $d_i$ from supported local Frosting thickness cues, part-level guidance, and within-part smoothness, subject to geometry-aware displacement bounds (\cref{sec:frosting_preprocessing}; Appendix \ref{app:frosting_defur_impl}).
\input{figures/part_annotation}
For each part $p$, we compute $T_p$ as the 75th percentile of its Frosting shell widths, converted to centimeters. The default target strand length is $\ell_p=\ell_p^{\mathrm{shape}}=T_p m_p$, where the fixed part-specific multiplier $m_p$ accounts for the difference between shell thickness and length along a curved or oblique strand. An optional VLM-assisted variant (V2) combines $60\%$ of this length with $40\%$ of a VLM length estimate, limiting the result to $35$--$135\%$ of the VLM estimate. Root density (roots per unit surface area) is assigned independently for each part, with zero density in non-fur regions.

\subsection{Defurring and Fur Initialization}
\label{sec:frosting_preprocessing}

Given calibrated multiview images and foreground masks, NeuS2~\citep{neus2} reconstructs the coat's outer surface $M_{\mathrm{outer}}$. We estimate a plausible defurred mesh $M_{\mathrm{root}}$ beneath it for strand attachment.

\paragraph{Defurring.}
Gaussian Frosting surrounds a base mesh with an adaptive layer of 3D Gaussians. Trained on the same images, its layer width provides local fur-thickness cues, not direct measurements of skin depth or strand length. We label body parts using ALIGN-Parts~\citep{alignparts2025}, with manual verification, avoiding SMAL fitting.
We transfer nearby shell widths using part labels and normal agreement, then calibrate them with coarse part-thickness references where available. Rather than copying the inner shell, we solve for smooth inward vertex displacements that balance supported local cues with part-level guidance. Weakly supported regions rely more on part-level estimates, while smoothing is weaker across part boundaries. We keep non-fur regions fixed, bound displacement near opposing surfaces, and reduce offsets causing face flips, severe collapse, or new self-intersections. We obtain $M_{\mathrm{root}}$ by moving each outer-mesh vertex $x_i$ to $x_i^{\mathrm{root}}=x_i-d_i n_i$, where $d_i$ is the estimated inward displacement and $n_i$ its outward unit normal.
\begin{figure*}[tb]
    \centering
    \includegraphics[width=\linewidth]{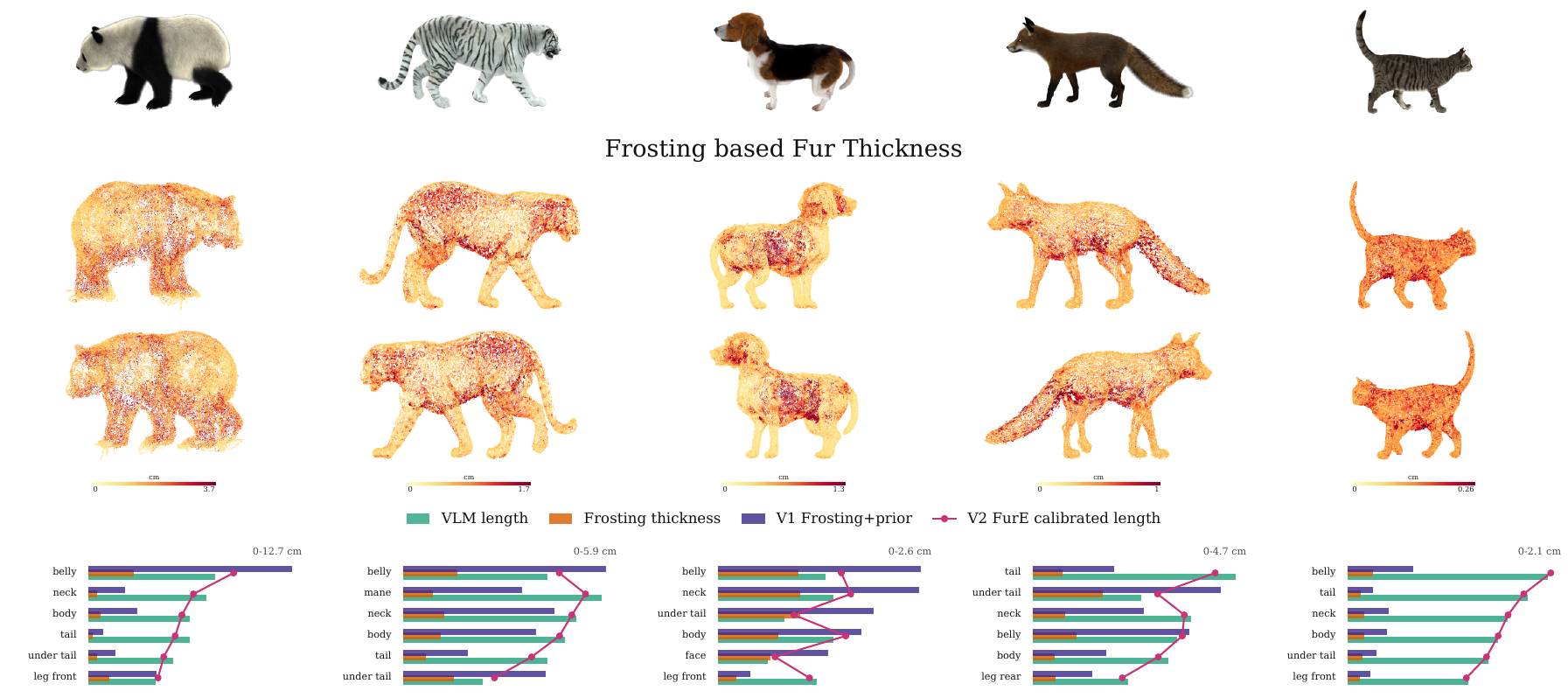}
    \caption{\textbf{Frosting thickness and strand-length calibration.}
    Middle rows: local Frosting shell thickness, with warmer colors indicating thicker shells. Bottom: per-part comparisons of raw Frosting thickness, VLM length estimates, and our V1/V2 strand lengths. Shell thickness provides relative geometric evidence, not a direct strand-length measurement.}
    \label{fig:fure-preprocess}
\end{figure*}

\begin{wrapfigure}{r}{0.51\textwidth}
    \centering
    \includegraphics[width=\linewidth]{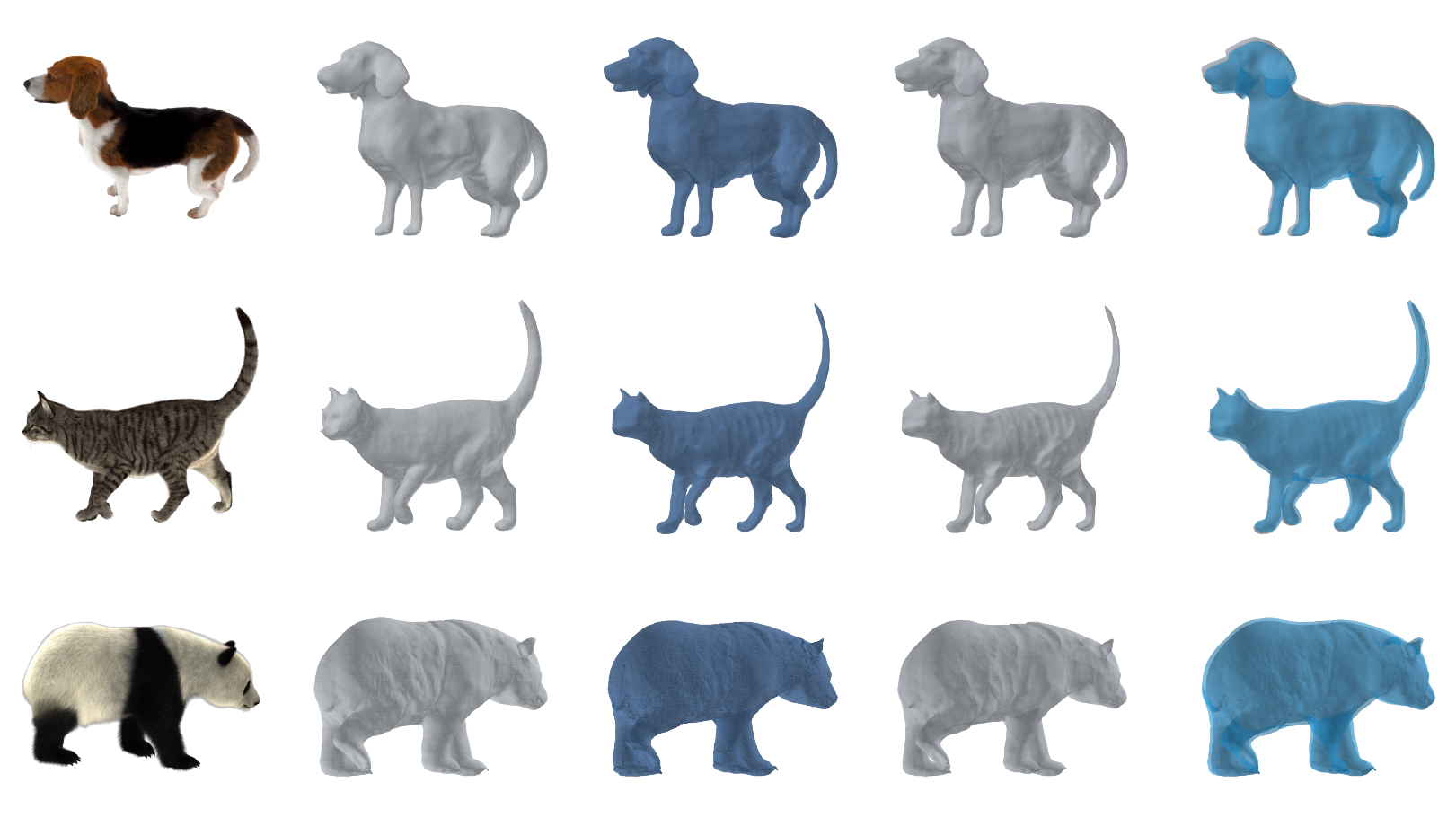}
    \caption{\textbf{Defurred geometry comparison.}
    Left to right: input model, $M_{\mathrm{outer}}$, our defurred mesh $M_{\mathrm{root}}$, NeuralFur's defurred mesh, and $M_{\mathrm{root}}$ overlaid on $M_{\mathrm{outer}}$.}
    \label{fig:furless}
    \vspace{-10pt}
\end{wrapfigure}

\paragraph{Fur initialization.}
Strand length is estimated separately from root displacement: curved or oblique strands can be longer than the coat is thick. For each fur-bearing part $p$, V1 initializes strand length as $\ell_p^{(1)}=m_pT_p$, where $T_p$ is the 75th percentile of its shell widths, converted to centimeters, and $m_p$ is a fixed part-specific multiplier. Optional V2 blends this length with a coarse VLM or supplied reference length. These variants affect length initialization, not defurring. We assign root-sampling densities independently for each part, with zero or near-zero density and zero strand length in non-fur regions. Full calibration, optimization, and initialization details are given in the appendix.

%% file: figures/method_fig.tex
\begin{figure*}[htb]
    \centering
    \includegraphics[width=\linewidth]{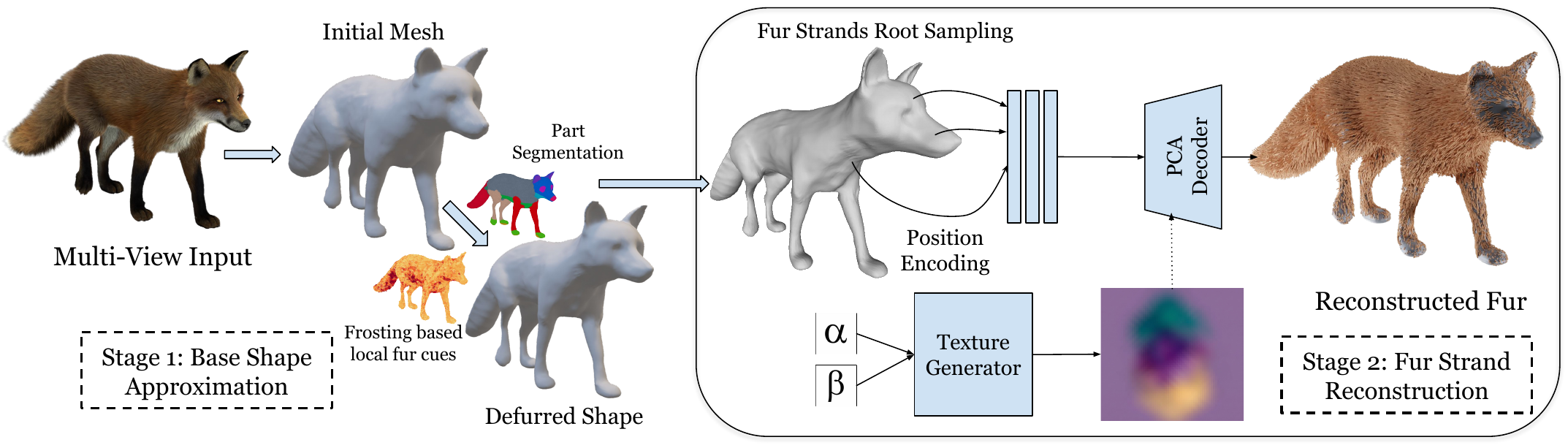}
    \caption{
    FurE has 2 stages, 1: Reconstructing the furless mesh geometry from multi view images using Gaussian Frosting by shrinking the initial reconstructed mesh with calculated fur length and 2: a PCA based strand decoder with an optional texture generator to generate 3D fur strands. Our method is optimized end to end from multi-view images in 3DGS compatible framework..}
    \label{fig:method-details}
\end{figure*}

%% file: figures/part_annotation.tex
\begin{wrapfigure}{r}{0.55\textwidth}
    \centering
    \includegraphics[width=\linewidth]{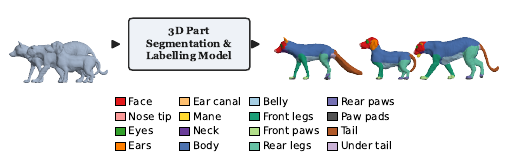}
    \caption{\textbf{3D Part Annotation.}
    We use 3D part segmentation and naming models like ~\cite{alignparts2025, 10.1145/1778765.1778839, partfield2025}
    instead of SMAL fitting, enabling fast 3D part segmentation even for non-quadruped mammals.}
    \label{fig:method-partseg}
    \vspace{-9pt}
\end{wrapfigure}

%% file: sec/3_method.tex
\subsection{Codec-based Strand Learning}
\label{strand_gen}

Rather than independently optimizing each strand's $L$ 3D points,
FurE learns compact shape coefficients. For each sampled root
$\mathbf{r}_i\in M_{\mathrm{root}}$, a learned function predicts
$\mathbf{z}_i=E_\theta(\gamma(\mathbf{r}_i),j,\rho(\mathbf{r}_i),\ell_j)$,
using positional encoding $\gamma$, part label $j=\pi(\mathbf{r}_i)$,
normalized local Frosting thickness $\rho$, and target length $\ell_j$.
A PCA decoder initialized from PERM's human-hair
basis~\citep{perm2025} produces a local strand
$\hat{S}_i=D_\psi(\mathbf{z}_i)$ with points $\hat{\mathbf{s}}_{ik}$.
We anchor, scale, and orient these points through
$\mathbf{s}_{ik}=\mathbf{r}_i+
\frac{\ell_j\,T_i(\hat{\mathbf{s}}_{ik}-\hat{\mathbf{s}}_{i1})}
{s_{\mathrm{cm}}(a_i+\epsilon)}$,
where $a_i$ is the sum of decoded segment lengths,
$T_i$ is the root's orthonormal tangent--bitangent--normal frame,
$s_{\mathrm{cm}}$ converts scene units to centimeters, and
$\epsilon>0$ prevents division by zero.

We show that human-hair priors can support animal fur reconstruction
despite differences in length, density, and growth direction.
Root sampling, local frames, and part-specific lengths control strand
placement, orientation, and scale. Reconstruction is instance-specific:
we optimize $\theta$ and fine-tune $\psi$ using the target animal's
multiview images (\cref{render_opt}), without a separate animal-fur
training dataset, while retaining explicit, editable strands.

Strand codes can also be represented as a 2D UV texture on
$M_{\mathrm{root}}$, with guide $G$ and style $S_{\mathrm{tex}}$ maps
for global structure and local strand detail. The texture generator
is optional.

\subsection{Rendering and Optimization}\label{render_opt}

Next, in order to integrate the reconstructed strands into the 3DGS differentiable rendering framework, we attach cylindrical Gaussians to the strands with lengths significantly larger than their diameters. Each line segment of a strand is represented
by a Gaussian whose length matches the segment length
and orientation aligns with the local tangent direction. Each
Gaussian is further associated with trainable spherical harmonic coefficients for appearance modeling, allowing the
photometric supervision to refine the geometric structure.

Each decoded strand is represented as a continuous chain of anisotropic Gaussian primitives, whose elongated shapes tightly follow the strand's local geometry. Specifically, we attach cylindrical Gaussians along the reconstructed strands
and integrate them into the 3DGS differentiable rendering
framework. Each line segment of a strand is represented
by a Gaussian whose length matches the segment length and orientation aligns with the local tangent direction.FurE optimizes the latent field parameters with multi-view losses:
\begin{equation}
\mathcal{L} =\lambda_{\mathrm{rgb}}\mathcal{L}_{\mathrm{rgb}}+
\lambda_{\mathrm{sil}}\mathcal{L}_{\mathrm{sil}}
+\lambda_{\mathrm{ori}}\mathcal{L}_{\mathrm{ori}}
+\lambda_{\mathrm{chm}}\mathcal{L}_{\mathrm{chm}}
+\lambda_{\mathrm{sdf}}\mathcal{L}_{\mathrm{sdf}} +\lambda_{\mathrm{mask}}\mathcal{L}_{\mathrm{mask}}
\end{equation}
$\mathcal{L}_{\mathrm{sil}}$ matches rendered masks, $\mathcal{L}_{\mathrm{ori}}$ matches image-space orientation $\mathcal{L}_{\mathrm{chm}}$ attracts fur to the outer envelope, $\mathcal{L}_{\mathrm{sdf}}$ prevents penetration into the mesh body and $\mathcal{L}_{\mathrm{mask}}$ is the loss between the GT and rendered fur mask.

%% file: sec/4_experiments.tex
\input{figures/main_comparison}

\section{Experiments}
\label{sec:experiments}
\begin{wraptable}{r}{0.4\textwidth}
    \centering
    \vspace{-10pt}
    \caption{\textbf{Runtime comparison.}
    \small Both methods use 36 views and an A5000 GPU.}
    \label{tab:fure-runtime}

    \small
    \setlength{\tabcolsep}{3pt}
    \renewcommand{\arraystretch}{1.0}
    \begin{tabular*}{\linewidth}{@{\extracolsep{\fill}}lcc@{}}
        \toprule
        Stage           & NeuralFur & \textbf{FurE} \\
        \midrule
        Strand training & 10.5\,h   & \textbf{52\,min} \\
        Preprocessing   & 10\,h     & \textbf{1\,h} \\
        \bottomrule
    \end{tabular*}
\end{wraptable}

\paragraph{Dataset}
We evaluate our method on five synthetic fur styles across different animals from the Artemis~\citep{10.1145/3528223.3530086} dataset, training on 36 uniformly sampled frames per sequence and evaluating novel-view rendering on the remaining frames. We further demonstrate results on a real-world bison sequence to demonstrate generalization beyond synthetic data. We compare our method against both surface reconstruction and strand-based reconstruction baselines.

\paragraph{Quantitative results}
\label{quant_results}
To quantitatively compare our method against GaussianHairCut and NeuralFur we use a synthetic tiger asset with artist generated ground truth strand based fur. As shown in Tab.~\ref{tab:thresholds_fure} we compute the precision,recall and F-score between   the ground truth and reconstructed strands. We further evaluate on the four synthetic scenes from Artemis~\citep{10.1145/3528223.3530086} using unsupervised geometric metrics that assess strand consistency in both local and global spaces, as well as proximity to the surface (Table.~\ref{tab:fure_quant_all}). The metrics capture three aspects of strand geometry: (1) length consistency, measured by the global mean $\mu_L$
and standard deviation $\sigma_L$ of strand lengths; (2) strand curvature, assessed via local and global curvature variance $\text{Var}_\text{loc}(\kappa)$
and $\text{Var}_\text{glob}(\kappa)$
and, (3) strand orientation, quantified by the local variance of strand directions $\text{Var}_\text{loc}(\text{dir})$
and, more finely, by $\text{Var}^\text{first}_\text{loc}(\text{dir})$, which restricts the direction estimate to the first segment of each strand. 
Following the evaluation protocol of GaussianHair, we also quantitatively assess the visual fidelity of our reconstructed fur strands by attaching strand-aligned Gaussians and rendering them, reporting PSNR, LPIPS, and SSIM metrics across four scenes in Table~\ref{tab:rendering_quant_all}, and find that FuE largely outperforms NeuralFur and Gaussian Haircut, especially considering the significant amount ($10\times$) of efficiency we bring about during training, as shown in ~\autoref{tab:fure-runtime}.

\paragraph{Qualitative results}
\label{quali_results}
We evaluate our approach against several baselines spanning animal reconstruction (GenZoo~\cite{niewiadomski2024generativezoo}),strand-based human hair modeling (Gaussian Haircut ~\cite{zakharov2024gaussianhaircut}), and strand-based animal fur modeling ~\cite{NeuralFur26}. As shown in Figure ~\ref{fig:comparison}, the neural surface methods produce only coarse outer geometry of the animal, lacking any strand-level detail in the reconstruction. Although Gaussian Haircut yields high-fidelity strand reconstructions for human hair, it struggles to generalize to fur geometry primarily because of absence of explicit part level strand length, which is implicitly optimized through photometric supervision. Our method, by contrast, recovers detailed, accurate fur structure directly from images during our defurring stage, without any of these limitations.

\begin{table}[ht]
\centering
\caption{Quantitative Evaluation on Synthetic Tiger with GT strands.}
\begin{tabular}{l ccc ccc ccc}
\toprule
 & \multicolumn{9}{c}{\textbf{Thresholds: cm / degrees}} \\
\cmidrule(lr){2-10}
\textbf{Method} & 2/20 & 3/30 & 4/40 & 2/20 & 3/30 & 4/40 & 2/20 & 3/30 & 4/40 \\
\cmidrule(lr){2-4} \cmidrule(lr){5-7} \cmidrule(lr){8-10}
 & \multicolumn{3}{c}{\textbf{Precision}} & \multicolumn{3}{c}{\textbf{Recall}} & \multicolumn{3}{c}{\textbf{F-score}} \\
\midrule
GaussianHairCut & 16.24 & 25.51 & 32.34 & 23.51 & 36.04 & 45.87 & 19.21 & 29.87 & 37.93 \\
NeuralFur & 26.22 & 39.32 & 48.05 & 20.58 & 34.08 & 45.69 & 23.06 & 36.51 & 46.84 \\
Ours & \textbf{27.62} & \textbf{41.78} & \textbf{51.20} & \textbf{21.69} & \textbf{38.00} & \textbf{50.52} & \textbf{24.30} & \textbf{39.80} & \textbf{50.86} \\
\bottomrule
\end{tabular}
\label{tab:thresholds_fure}
\end{table}

\begin{table}[t]
    \centering
    \caption{Unsupervised geometry consistency metrics for length, direction, and curvature across four scenes (Panda, Fox, Cat, whiteTiger), evaluated for both local and global cases. Note that there are no ground truth strands available for these metrics.}
    \label{tab:fure_quant_all}

    \small
    \setlength{\tabcolsep}{2pt}
    \renewcommand{\arraystretch}{1.0}
    \sisetup{
        round-mode=none,
        group-digits=false
    }

    \begin{tabular*}{\linewidth}{
        @{\extracolsep{\fill}}
        c l
        >{\columncolor[gray]{0.95}}S[table-format=1.6]
        *{6}{S[table-format=1.6]}
        @{}
    }
        \toprule
        & Metric
        & {\textbf{Ours}}
        & {\shortstack{Gaussian\\HairCut}}
        & {NeuralFur}
        & {\shortstack{Neural\\HairCut}}
        & {\shortstack{Fixed\\Length}}
        & {Defurring}
        & {ShapePrior} \\
        \midrule

        \multirow{6}{*}{\rotatebox[origin=c]{90}{\textbf{Panda}}}
        & $\mu_L$
        & 5.22 & 8.79 & 5.23 & 8.90 & 6.00 & 5.19 & 5.24 \\
        & $\sigma_L$
        & 1.38 & 2.04 & 1.38 & 2.04 & 0.00 & 1.38 & 1.39 \\
        & $\mathrm{Var}_{\mathrm{global}}$
        & 0.0004 & 0.0024 & 0.0004 & 0.0023 & 0.0002 & 0.0020 & 0.0001 \\
        & $\mathrm{Var}_{\mathrm{loc}}$
        & 0.000025 & 0.000118 & 0.000041 & 0.000118 & 0.000050 & 0.00020 & 0.00006 \\
        & $\mathrm{Var}_{\mathrm{loc}}(\mathrm{dir})$
        & 0.046 & 0.56 & 0.043 & 0.50 & 0.054 & 0.068 & 0.051 \\
        & $\mathrm{Var}_{\mathrm{loc}}^{\mathrm{first}}(\mathrm{dir})$
        & 0.011 & 0.19 & 0.081 & 0.19 & 0.080 & 0.023 & 0.013 \\
        \midrule

        \multirow{6}{*}{\rotatebox[origin=c]{90}{\textbf{Fox}}}
        & $\mu_L$
        & 3.80 & 7.190 & 3.00 & 7.16 & 3.00 & 3.82 & 4.00 \\
        & $\sigma_L$
        & 1.15 & 3.18 & 1.18 & 3.18 & 0.00 & 1.14 & 1.20 \\
        & $\mathrm{Var}_{\mathrm{global}}$
        & 0.00036 & 0.006 & 0.00050 & 0.007 & 0.00040 & 0.001 & 0.00090 \\
        & $\mathrm{Var}_{\mathrm{loc}}$
        & 0.000016 & 0.0056 & 0.00018 & 0.0056 & 0.00089 & 0.00026 & 0.000025 \\
        & $\mathrm{Var}_{\mathrm{loc}}(\mathrm{dir})$
        & 0.094 & 0.46 & 0.086 & 0.50 & 0.072 & 0.60 & 0.011 \\
        & $\mathrm{Var}_{\mathrm{loc}}^{\mathrm{first}}(\mathrm{dir})$
        & 0.017 & 0.47 & 0.019 & 0.47 & 0.010 & 0.14 & 0.016 \\
        \midrule

        \multirow{6}{*}{\rotatebox[origin=c]{90}{\textbf{Cat}}}
        & $\mu_L$
        & 1.60 & 4.94 & 1.65 & 5.00 & 1.70 & 1.65 & 1.70 \\
        & $\sigma_L$
        & 0.78 & 2.17 & 0.53 & 0.31 & 0.00 & 0.78 & 0.90 \\
        & $\mathrm{Var}_{\mathrm{global}}$
        & 0.00025 & 0.002 & 0.00060 & 0.002 & 0.0003 & 0.0014 & 0.0008 \\
        & $\mathrm{Var}_{\mathrm{loc}}$
        & 0.00001 & 0.01 & 0.000085 & 0.01 & 0.00006 & 0.0002 & 0.000014 \\
        & $\mathrm{Var}_{\mathrm{loc}}(\mathrm{dir})$
        & 0.085 & 0.58 & 0.036 & 0.58 & 0.072 & 0.095 & 0.010 \\
        & $\mathrm{Var}_{\mathrm{loc}}^{\mathrm{first}}(\mathrm{dir})$
        & 0.013 & 0.56 & 0.010 & 0.56 & 0.08 & 0.032 & 0.014 \\
        \midrule

        \multirow{6}{*}{\rotatebox[origin=c]{90}{\textbf{whiteTiger}}}
        & $\mu_L$
        & 3.70 & 6.20 & 4.00 & 6.00 & 5.00 & 3.71 & 3.80 \\
        & $\sigma_L$
        & 1.60 & 0.31 & 1.80 & 0.31 & 0.00 & 1.60 & 1.60 \\
        & $\mathrm{Var}_{\mathrm{global}}$
        & 0.0003 & 0.0005 & 0.0003 & 0.0008 & 0.0025 & 0.0005 & 0.00027 \\
        & $\mathrm{Var}_{\mathrm{loc}}$
        & 0.000025 & 0.0019 & 0.000018 & 0.000022 & 0.00033 & 0.00006 & 0.000025 \\
        & $\mathrm{Var}_{\mathrm{loc}}(\mathrm{dir})$
        & 0.04 & 0.50 & 0.05 & 0.00019 & 0.09 & 0.05 & 0.04 \\
        & $\mathrm{Var}_{\mathrm{loc}}^{\mathrm{first}}(\mathrm{dir})$
        & 0.008 & 0.46 & 0.013 & 0.41 & 0.08 & 0.08 & 0.008 \\
        \bottomrule
    \end{tabular*}
\end{table}

\begin{table}[htb]
    \centering
    \caption{Quantitative comparison of rendering quality metrics across four scenes.}
    \label{tab:rendering_quant_all}

    \small
    \setlength{\tabcolsep}{3pt}
    \renewcommand{\arraystretch}{1.05}

    \begin{tabular*}{\linewidth}{@{\extracolsep{\fill}} l *{9}{c} @{}}
        \toprule
        & \multicolumn{3}{c}{GaussianHairCut}
        & \multicolumn{3}{c}{NeuralFur}
        & \multicolumn{3}{c}{\textbf{FurE (Ours)}} \\
        \cmidrule(lr){2-4}
        \cmidrule(lr){5-7}
        \cmidrule(l){8-10}
        Scene
        & PSNR$\uparrow$ & LPIPS$\downarrow$ & SSIM$\uparrow$
        & PSNR$\uparrow$ & LPIPS$\downarrow$ & SSIM$\uparrow$
        & PSNR$\uparrow$ & LPIPS$\downarrow$ & SSIM$\uparrow$ \\
        \midrule
        Panda
        & 43.82 & 0.3281 & 0.6130
        & \textbf{43.91} & 0.3266 & 0.6135
        & \textbf{43.91} & \textbf{0.3258} & \textbf{0.6144} \\
        whiteTiger
        & 43.09 & 0.3128 & 0.6169
        & \textbf{43.11} & 0.3107 & 0.6181
        & 43.10 & \textbf{0.3097} & \textbf{0.6187} \\
        Fox
        & 44.86 & 0.3284 & 0.6082
        & 44.87 & 0.3268 & 0.6090
        & \textbf{44.88} & \textbf{0.3261} & \textbf{0.6092} \\
        Cat
        & 49.82 & 0.3141 & 0.6141
        & 49.82 & 0.3125 & 0.6132
        & \textbf{49.84} & \textbf{0.3120} & \textbf{0.6165} \\
        \bottomrule
    \end{tabular*}
\end{table}

\textbf{Defurring Ablation}
Figure~\ref{fig:fure-preprocess} visualizes the Frosting-based cues used by FurE. As can be observed, Frosting provides a local, view-consistent shell-width cue indicating where the reconstructed surface has thick fur. Since radial shell thickness is only a lower bound on strand arc length, we calibrate it with part-level priors. As can be observed in ~\autoref{fig:comparison}, our fur length and width calculations and methodology is validated by the visual results. Figure~\ref{fig:furless} also visually validates our defurring method, as we observe that we produce results similar to NeuralFur while being significantly faster.
\input{figures/main_ablation}

\paragraph{Applications}
Once the fur strands are reconstructed, they can be directly imported into industry-standard game engines such as Blender and Unreal Engine, enabling seamless integration into real-time rendering and animation pipelines. We further demonstrate in \autoref{fig:fox_sim_wind} the physical plausibility of the reconstructed strands by simulating strong wind dynamics, showing that the fur responds with realistic, physically coherent motion.

%% file: figures/main_comparison.tex
\begin{figure*}[!ht]
    \centering
    \includegraphics[width=\linewidth]{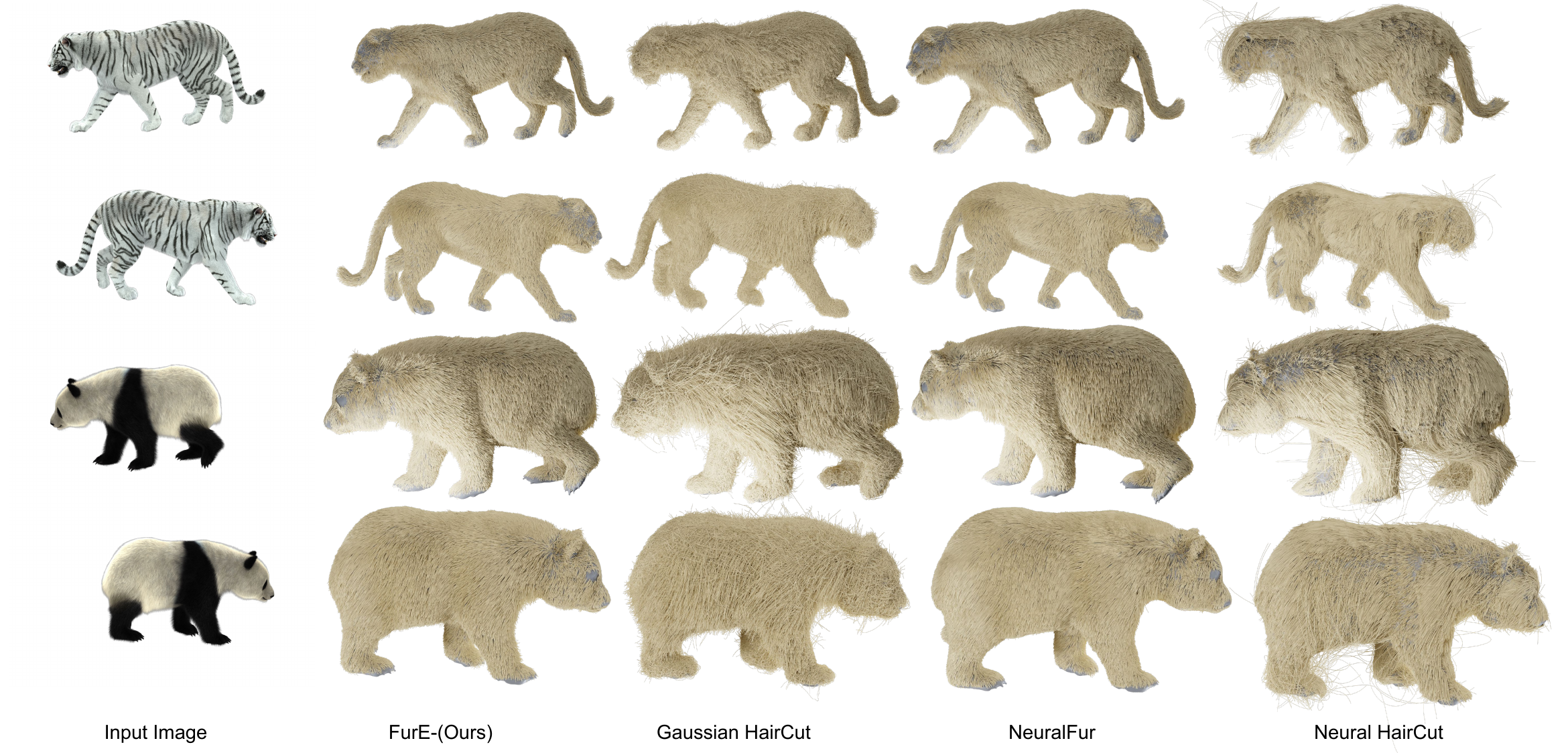}
    \includegraphics[width=\linewidth]{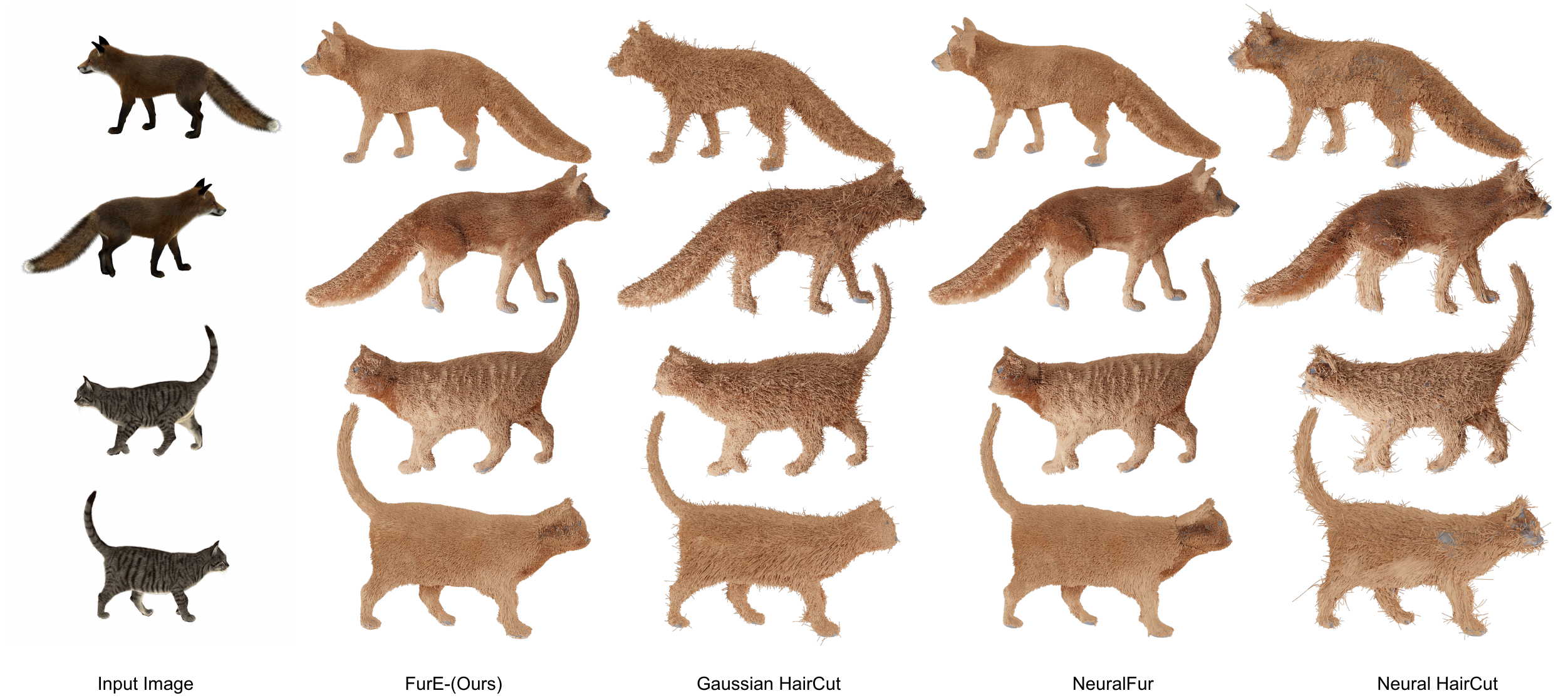}
    \caption{
    We show the qualitative comparisons between our method and existing baselines. Surface reconstruction approaches yield overly coarse and inaccuarte geometry, failing to capture fine strand-level detail. Adapting Gaussian Haircut  hair reconstruction method, results in inconsistent strand lengths and visible reconstruction artifacts. While Neural Fur and our method, produces accurate and coherent strand-based geometry across all evaluated subjects, we do it 10x faster.
    }
    \label{fig:comparison}
\end{figure*}
\begin{figure*}[!htb]
    \centering
    \includegraphics[width=\linewidth]{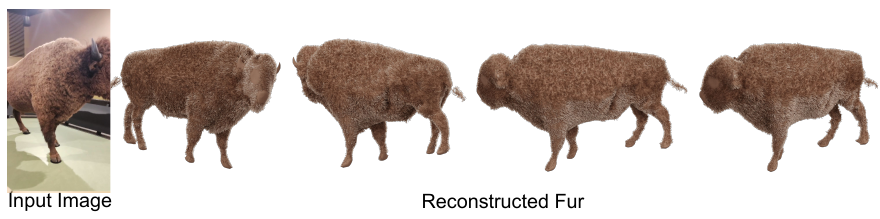}
    \caption{\textbf{Instance-specific real-world fur reconstruction.}
    FurE reconstructs explicit, editable strands from noisy real world multiview
    images of a bison. NeuralFur~\citep{NeuralFur26} struggles on this
    sequence (discussed in the appendix). We also release the associated
 data.}
    \label{fig:bison_results}
\end{figure*}

%% file: figures/main_ablation.tex
\begin{figure*}[htb]
    \label{ablation_fig}
    \centering
    \includegraphics[width=\linewidth, trim=0 0 0 0cm, clip]{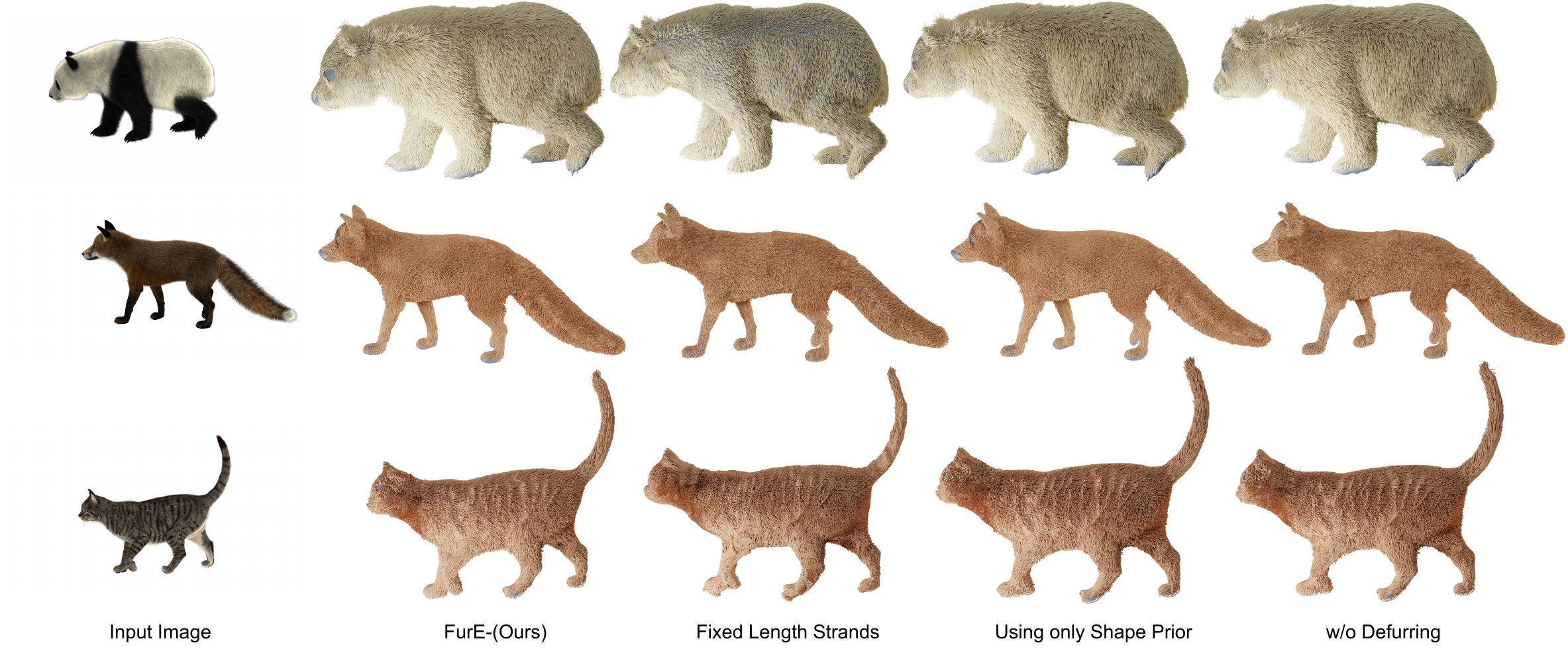}
    \caption{
   We show the qualitative evaluation of key design choices in our framework, including strand length parameterization per semantic body region, fur geometry representation, shape prior integration, and the role of the defurring stage in enabling accurate geometry reconstruction.}
    \label{fig:ablation_results}
\end{figure*}

\begin{figure*}[htb]
    \centering
    \includegraphics[width=\linewidth]{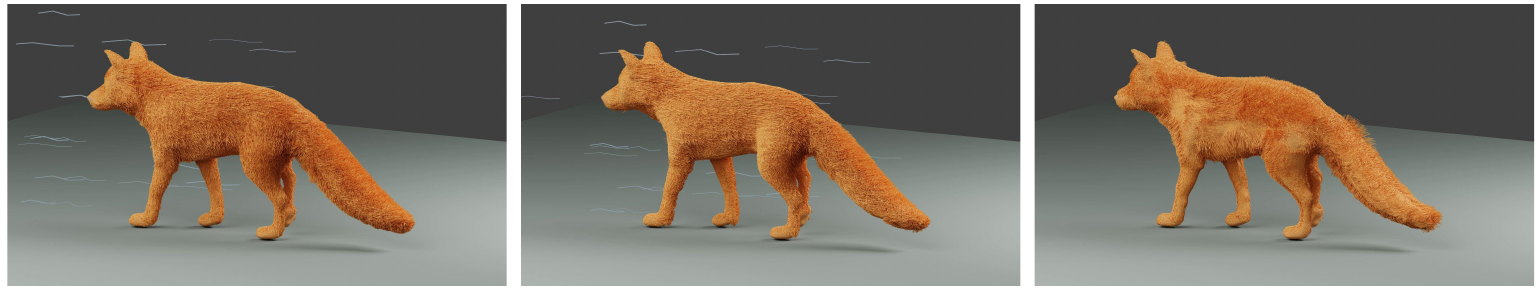}
\caption{Reconstructed fur strands imported into Blender subjected to strong wind simulation, demonstrating physically plausible dynamic behavior. See video in supplementary.}
    \label{fig:fox_sim_wind}
\end{figure*}

\paragraph{Strand Generation Ablations}
We conduct comprehensive ablation studies to assess the contribution of each component in our strand generation pipeline. Specifically, we examine three configurations: (1) using a fixed strand length uniformly across all semantic body parts, (2) estimating strand length using only the shape prior without a defurred mesh, and (3) our full method (results in Figure~\ref{fig:ablation_results}). Assigning a fixed length across body regions leads to inaccurate fur reconstruction, particularly on the body and belly where strands are naturally longer. While removing the defurred mesh yields visually similar results, the quantitative results show inconsistency in direction and curvature.

%% file: sec/5_conclusion.tex
\subsection{More Visualizations}
\begin{figure}[!ht]
\centering
\includegraphics[width=\linewidth]{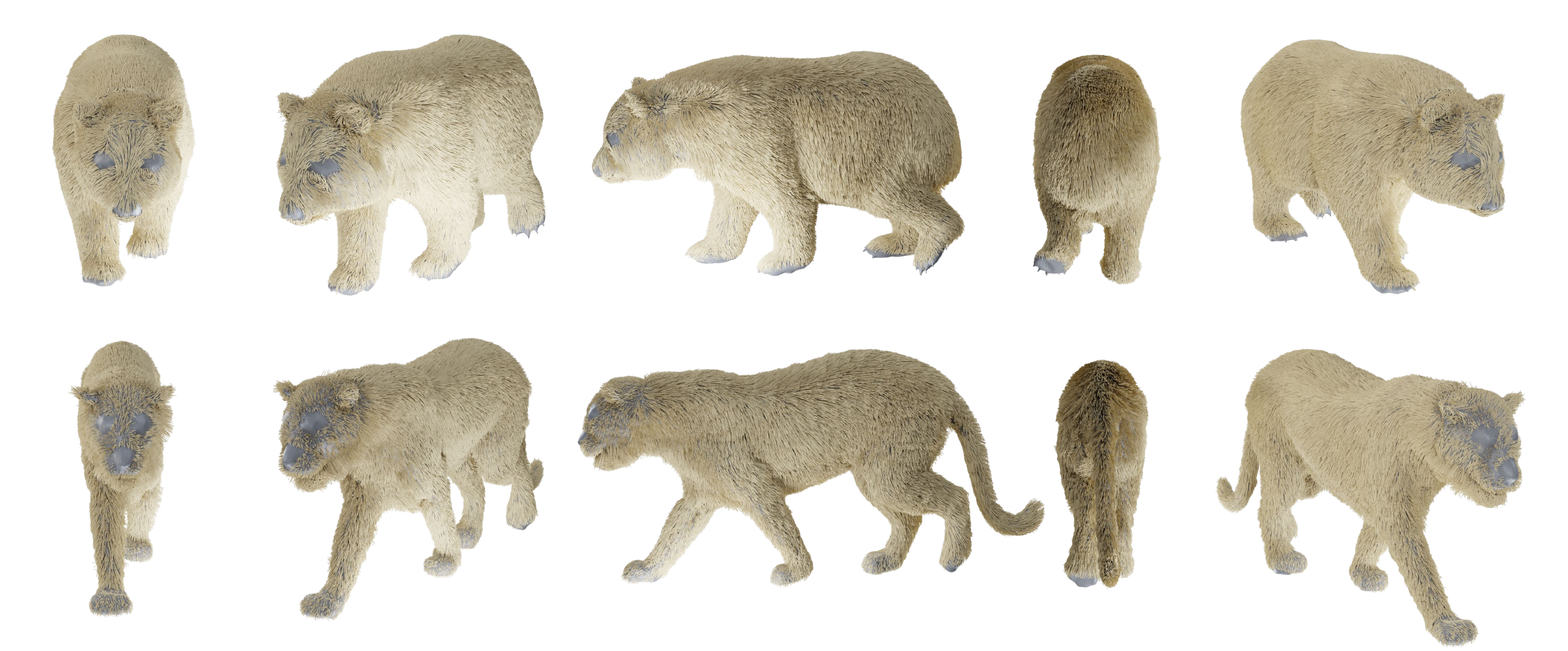}
\includegraphics[width=\linewidth]{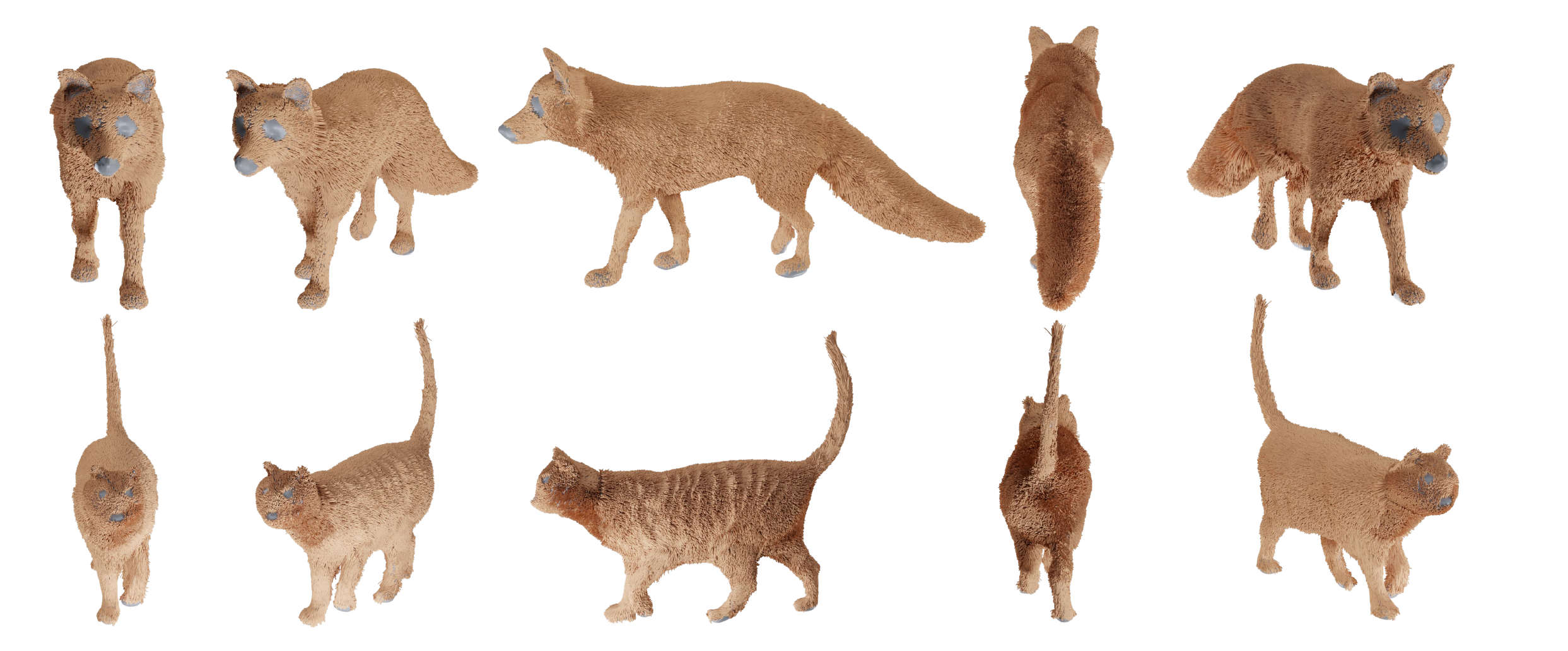}
\caption{\textbf{Rendering examples.} Renderings of reconstructed strand-based fur across different views of \emph{panda}, and \emph{white tiger}, \emph{fox}, \emph{cat},}
\label{fig:extra_comparison}
\end{figure}

In Figure \ref{fig:extra_comparison}, we provide additional novel view renderings of our reconstructed strand-based fur across the different animal subjects, including a panda, white tiger, fox, and cat. These qualitative results demonstrate FurE's ability to consistently capture instance specific fur characteristics, across different species of animals. Despite bypassing dense per-strand optimization, our codec-based approach maintains high-fidelity geometric details from multiple viewpoints.

\section{Conclusion}
\label{sec:conclusion}

We presented FurE for instance-specific reconstruction of explicit,
editable animal fur from calibrated multiview images. FurE adapts a
human-hair PCA basis to replace dense strand optimization with compact
code learning, without requiring a separate animal-fur training dataset.
Local Gaussian Frosting cues and part-aware priors guide defurring,
while direct part segmentation removes the need for SMAL fitting.
On Artemis, strand training takes under one hour, achieving a
$10\times$ speedup over NeuralFur with comparable rendering quality.
We further reconstruct fur from a noisy real-world bison sequence
on which NeuralFur fails. These results demonstrate efficient animal
fur reconstruction while retaining explicit strands for downstream
editing, rendering, and simulation.

However, several limitations still remain. Our evaluation lacks ground-truth cases
with strong local variation within body parts, such as shaved patches,
injuries, or irregular grooming. Such cases would better test local
Frosting cues against part-level priors. FurE also represents fur as
a single strand layer on one root surface, which is estimated before
strand optimization. Modeling multilayer coats and jointly optimizing
root placement and strand geometry under multiview supervision are
important next steps toward more automatic reconstruction.

%% file: sec/appendix.tex
\section{Appendix}
\subsection{Implementation Details for Frosting-Calibrated Defurring}
\label{app:frosting_defur_impl}

FurE estimates the strand-root surface and initializes strand lengths
separately. Root displacement determines where strands attach beneath the
coat; strand length determines their extent along the curve. Gaussian
Frosting supplies local thickness cues, while part-level references
calibrate these cues and provide guidance where local support is weak.
Shell width is not treated as a direct measurement of skin depth or
strand length.

\paragraph{Inputs and shell measurements.}
We reconstruct the visible furry mesh $M_{\mathrm{outer}}$ using
NeuS2~\citep{neus2}. Each vertex $x_i$ has an outward unit normal $n_i$
and a body-part label $p(i)$, obtained using
ALIGN-Parts~\citep{alignparts2025} with manual verification.

Gaussian Frosting~\citep{guedon2024gaussianfrosting} is trained on the same
images and expressed in the same coordinate frame. Let $y_k$, $\nu_k$,
and $w_k$ denote a shell vertex, its normal, and its stored shell width.
Shell vertices receive the
label of their nearest annotated outer-mesh vertex.

We retain an unrestricted nearest-shell width for each outer-mesh vertex:
\[
    k_0(i)=\arg\min_k\|x_i-y_k\|_2,
    \qquad
    \tau_i=w_{k_0(i)}.
\]
These raw widths supply the part statistics and length initialization.
For local defurring cues, we additionally check correspondence quality
as described below.

\paragraph{Supported local thickness.}
For each $x_i$, we consider up to 12 nearest shell vertices. A
correspondence is accepted only when its distance is below $0.03D$,
its part label matches $p(i)$, its normal satisfies
$n_i^\top\nu_k>0.3$, and its stored width passes the reliability threshold.
Here $D$ is the scene length scale used by the preprocessing.
The reliability threshold accepts widths at or below the global
97.5th percentile.

Let $\mathcal A_i$ contain the accepted correspondences and
$\omega_{ik}$ their distance/normal weights before normalization.
For vertices with positive support, we compute
\[
    \bar{\tau}_i
    =
    \frac{\sum_{k\in\mathcal A_i}\omega_{ik}w_k}
         {\sum_{k\in\mathcal A_i}\omega_{ik}},
    \qquad
    c_i=\max_{k\in\mathcal A_i}\omega_{ik}.
\]
Thus, $\bar{\tau}_i$ averages the supported widths, while $c_i$
measures correspondence support. Unsupported vertices receive $c_i=0$,
and their local term is omitted from the displacement objective.
Missing support is not interpreted as absence of fur.

\paragraph{Part-level calibration.}
Within each part, we cap the raw nearest-shell widths $\tau_i$ at
that part's 95th percentile. Let $W_{75,p}$ and $W_{95,p}$ denote
the 75th and 95th percentiles of these capped values. We combine
the shell statistic with a coarse part-thickness reference $h_p$:
\[
    B_p
    =
    \min\left(
        0.65\,W_{75,p}+0.35\,h_p,\;
        W_{95,p}
    \right).
\]
All quantities in this calibration use scene units. When a reference
is missing, we set $h_p=W_{75,p}$. An explicitly supplied zero remains
zero and is not treated as missing.

\paragraph{Bounded defurring.}
We estimate an inward displacement $d_i$ for each vertex by solving
\begin{equation}
\begin{aligned}
\min_{\mathbf d}\quad&
\sum_i c_i(d_i-t_i)^2
+\eta\sum_i(d_i-b_i)^2
+\lambda\sum_{(i,j)\in\mathcal E}a_{ij}(d_i-d_j)^2,\\
\text{s.t.}\quad&
0\leq d_i\leq u_i,\qquad
d_i=0\quad(i\in\mathcal B).
\end{aligned}
\label{eq:app_root_displacement}
\end{equation}
The first term follows supported local evidence, the second retains
part-level guidance, and the third smooths displacement across mesh
edges $\mathcal E$. We use $\eta=0.25$ and $\lambda=8$, with
inverse-edge-length weights:
\[
    a_{ij}
    =
    \frac{1}{\|x_i-x_j\|_2}
    \begin{cases}
        1, & p(i)=p(j),\\
        0.15, & p(i)\neq p(j).
    \end{cases}
\]

The protected set $\mathcal B$ fixes non-fur regions such as eyes,
nose tip, paw pads, and horns. To restrict inward movement near
opposing surfaces, we use
\[
    u_i=\min\left(0.45\,r_i^{\mathrm{hit}},\,0.05D\right),
\]
where $r_i^{\mathrm{hit}}$ is the inward-ray hit distance.
Missing hits use $r_i^{\mathrm{hit}}=0.1D$.
Here $u_i$ is a displacement bound.

We solve with L-BFGS-B,
for at most 2,500 iterations. The settings are
\texttt{ftol=1e-13}, \texttt{gtol=1e-7}, and \texttt{maxcor=8}.
We further reduce displacements that cause face inversions, severe
collapse, or new self-intersections. Using $d_i$ for the final checked
displacement, we form the root mesh through
\begin{equation}
    x_i^{\mathrm{root}}=x_i-d_i n_i.
    \label{eq:app_root_vertex}
\end{equation}
This estimates a plausible attachment surface beneath the coat. 

\paragraph{Metric scale and length statistics.}
Length initialization uses the raw, unrestricted nearest-shell widths
$\tau_i$, not the correspondence-filtered widths, capped defurring
statistics, or achieved displacements. For each fur-bearing part,
we compute the vertex-wise statistic
\[
    T_p
    =
    s_{\mathrm{cm}}\,
    \operatorname{P}_{75}\{\tau_i:p(i)=p\}.
\]
The scale factor is
\[
    s_{\mathrm{cm}}
    =
    \frac{\text{assumed eye separation in centimeters}}
         {\text{reconstructed eye separation in scene units}}.
\]
Centimeter-valued initialization requires this metric reference;
the current procedure has no validated automatic fallback when
metric information is unavailable. Non-fur parts receive $T_p=0$.

\paragraph{V1 and V2 length initialization.}
V1 initializes strand length using a fixed part-specific multiplier:
\[
    \ell_p^{\mathrm{shape}}=m_pT_p,
    \qquad
    \ell_p^{(1)}=\ell_p^{\mathrm{shape}}.
\]
The multiplier $m_p$ accounts for the difference between shell
thickness and length along a curved or oblique strand.
V1 does not use VLM length calibration, but still uses the metric
scale and part-specific multipliers.

When a coarse VLM or supplied reference length
$\ell_p^{\mathrm{VLM}}$ is available, V2 uses
\[
    \ell_p^{(2)}
    =
    \operatorname{clip}\left(
        0.60\,\ell_p^{\mathrm{shape}}
        +0.40\,\ell_p^{\mathrm{VLM}},
        \;
        0.35\,\ell_p^{\mathrm{VLM}},
        \;
        1.35\,\ell_p^{\mathrm{VLM}}
    \right).
\]
Both lengths are expressed in centimeters. The clipping operation
limits the blended length to the stated interval. Without a reference,
V2 uses $\ell_p^{(2)}=\ell_p^{\mathrm{shape}}$.

The selected variant supplies the exported length $\ell_p$.
Non-fur parts receive zero length in both variants.
V1 and V2 change strand-length initialization, not the defurring
objective.

\paragraph{Strand width and root sampling.}
The strand-rendering width parameter is initialized as
\[
    W_p^{\mathrm{strand}}
    =
    \operatorname{clip}
    \left(0.03\,\ell_p,\;0.02,\;0.12\right).
\]
This parameter is separate from shell thickness and root displacement.
Each part also receives an independent base root density, measured
in roots per unit surface area. Root sampling uses the part labels
and local Frosting cues, with zero or near-zero density in non-fur
regions.

\paragraph{Orientation and connection to strand learning.}
Directional computes a face-based degree-2 power field with a
directional constraint on face 0. We normalize the resulting
directions and align their signs through a breadth-first traversal.
A bank of 180 Gabor orientations provides image-space fur-direction
cues.

Strand learning samples attachment points
$\mathbf r_s\in M_{\mathrm{root}}$ and uses their local frames,
part labels, and initialized lengths and widths. These sampled roots
are distinct from the displaced mesh vertices
$x_i^{\mathrm{root}}$. A root on part $p(\mathbf r_s)$ receives
the corresponding length $\ell_{p(\mathbf r_s)}$.

When used by the codec, $\rho(\mathbf r_s)$ denotes the normalized
local shell-width feature interpolated to the root. It is distinct
from correspondence support $c_i$ and root displacement $d_i$.

\begin{figure*}[htb]
    \includegraphics[width=\linewidth]{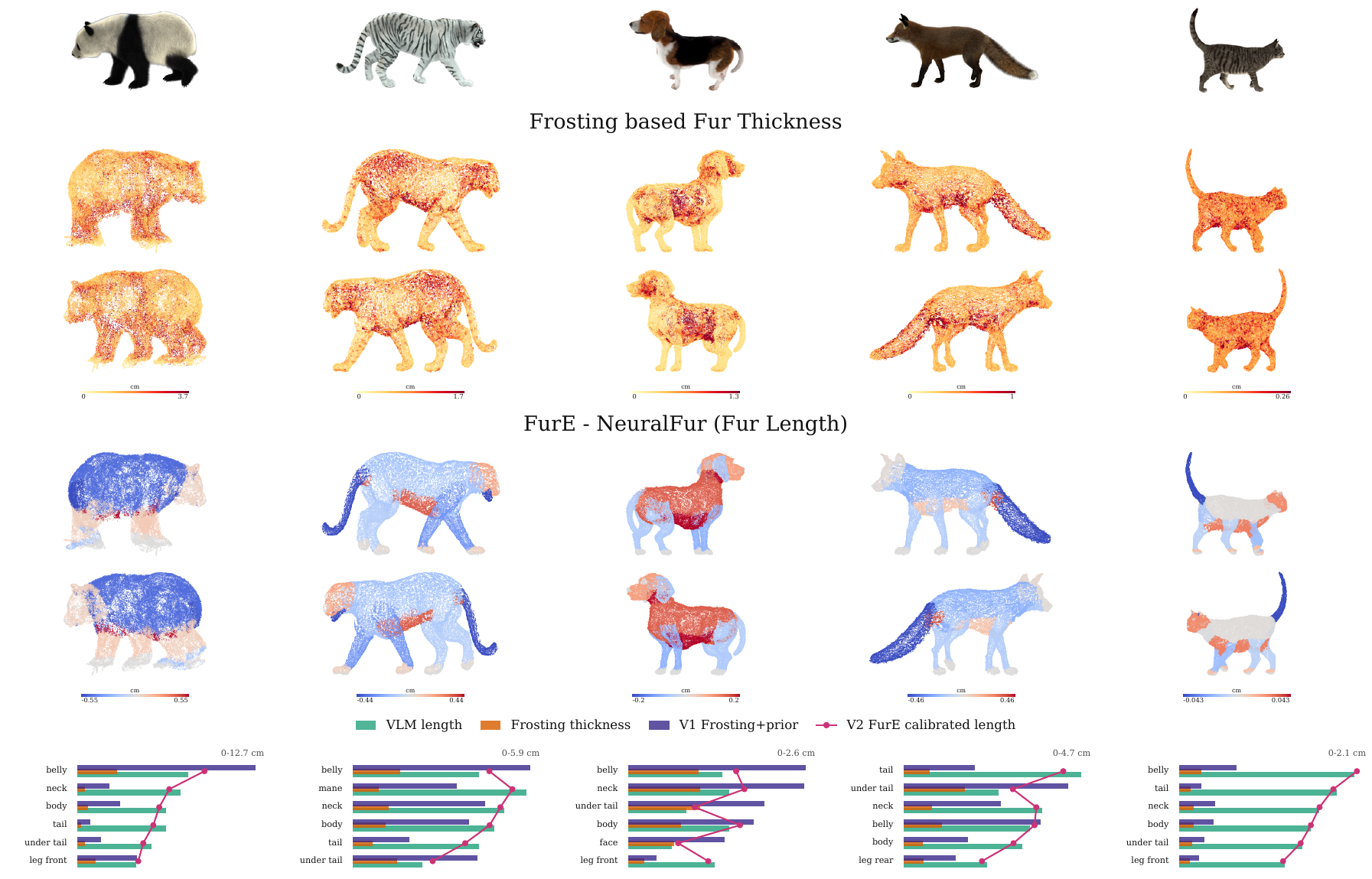}
    \caption{
    \textbf{Frosting thickness evidence and strand-length calibration.}
For each animal, the top row shows the original colored input rendering. The middle rows visualize the local Frosting shell thickness, where warmer colors indicate larger estimated fur-bearing thickness. The lower rows compare our calibrated fur-length estimate against the NeuralFur/VLM prior; blue regions are shorter than the prior and red regions are longer. The bottom charts summarize per-part values: raw Frosting thickness provides local geometric evidence, while the final calibrated length combines this signal with shape/part priors to produce NeuralFur-compatible strand lengths. Raw Frosting thickness is used as a relative local cue, not as a direct hair-length measurement.
    }
    \label{fig:fure-preprocess-appendix}
\end{figure*}

\subsection{Codec Strand Learning Details}
\label{app:codec}

FurE optimizes a compact strand code rather than dense per-strand control points.
For each sampled root $\mathbf{r}_i$, a root-conditioned encoder predicts a latent code $\mathbf{z}_i$, and a decoder maps this code to a normalized strand curve.
The root tangent frame, part label, Frosting density, and initialized strand length scale the decoded curve into world space. We use a continous latent texture for strand optimization. We train with 15K strand for every iteration for 2500 iterations and finetune the PCA  based decoder. We use the same hyper-parameters as NeuralFur.  
\begin{figure*}[htb]
    \centering
    \includegraphics[width=\linewidth]{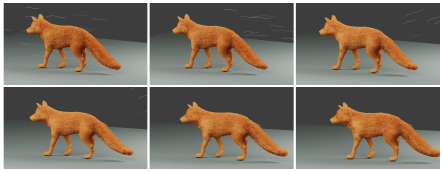}
    \vspace{-0.5cm}
\caption{Reconstructed fur strands imported into Blender subjected to strong wind simulation, demonstrating physically plausible dynamic behavior}
    \label{fig:more_fox_sim_wind}
\end{figure*}

\subsection{Applications}
As we demonstrate in the main and in Fig.~\ref{fig:more_fox_sim_wind}, the fur strands reconstructed by FurE can be directly imported into commonly used standard 3D game engines, such as Blender and Unreal Engine. The structural plausibility of the generated strands enables downstream physics simulations. For example, subjecting the imported strands to simulated wind dynamics produces realistic and coherent motion.

\subsection{Limitations and Future Work}
While FurE enables efficient, fur reconstruction, several limitations still remain. First, our current framework requires calibrated multi-view images of static animal. Since capturing dense, static multi-view images of living animals is often very difficult, extending FurE to handle dynamic motion is the next step. Additionally our defurring process relies on initial Gaussian Frosting cues and dense camera viewpoints. Severe self-occlusions in the input images and the animal fur  can degrade the shell thickness and the fur length estimate and hence the need for stronger geometric priors for  future iterations.

\begin{figure}[!ht]
    \centering
    \includegraphics[width=\linewidth]{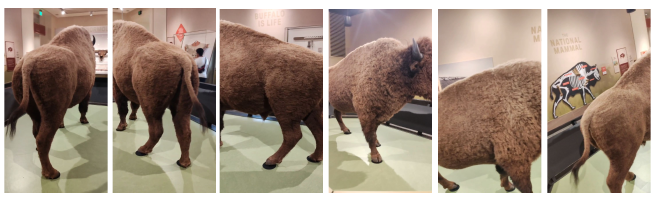}
    \caption{\textbf{Additional real-world samples.} These are more samples from the real-world use case. FurE successfully reconstructs explicit, editable strands from noisy multiview images of a bison. We also release the associated data.}
    \label{fig:more_bison_inputs}
\end{figure}

\subsection{Real World Fur Reconstruction}
To demonstrate the generalizability and robustness of our approach , we evaluate FurE on a real-world, instance-specific bison sequence. As shown in Figure ~\ref{fig:bison_results} , FurE successfully handles the real world noisy multi-view images to reconstruct the strand-based geometry. NeuralFur on the other hand fails to capture the underlying geometric surface for the bison. This surface degradation prevents it from reconstructing coherent and plausible fur geometry. We also release the associated multiview data to encourage future work on fur reconstruction on real world samples.